\documentclass[utf8x]{article-hermes_englb}
\usepackage[labelfont=bf,textfont=it,labelsep=period,justification=raggedright,singlelinecheck=false]{caption}
\usepackage{xcolor}
\usepackage{url}
\usepackage{booktabs}
\usepackage{multirow}
\usepackage{subcaption}

\submitted[15/11/2003]{TAL~64-03}{1}

\title[Explanation sensitivity to randomness]{Explanation sensitivity to the randomness of large language models: the case of journalistic text classification\vspace*{-0.7cm}}

\author{Jérémie Bogaert$^1$, Marie-Catherine de Marneffe$^2$, \\ Antonin Descampe$^2$, Louis Escouflaire$^2$, \\ Cédrick Fairon$^2$, François-Xavier Standaert$^1$} 

\address{$^1$ UCLouvain, ICTEAM Institute, Louvain-la-Neuve, Belgium\\
$^2$ UCLouvain, ILC Institute, Louvain-la-Neuve, Belgium\\\\
e-mails: firstname.lastname@uclouvain.be}

\abstract{Large language models (LLMs) perform very well in several natural language processing tasks but raise explainability challenges. In this paper, we examine the effect of random elements in the training of LLMs on the explainability of their predictions. We do so on a task of opinionated journalistic text classification in French. Using a fine-tuned CamemBERT model and an explanation method based on relevance propagation, we find that training with different random seeds produces models with similar accuracy but variable explanations. We therefore claim that characterizing the explanations’ statistical distribution is needed for the explainability of LLMs. We then explore a simpler model based on textual features which offers stable explanations but is less accurate. Hence, this simpler model corresponds to a different tradeoff between accuracy and explainability. We show that it can be improved by inserting features derived from CamemBERT's explanations. We finally discuss new research directions suggested by our results, in particular regarding the origin of the sensitivity observed in the training randomness.}


\begin{document}

\noindent\textit{This paper is a faithful translation of a paper which was peer-reviewed and published in the French journal} Traitement Automatique des Langues:

\smallskip

\noindent{Bogaert, J., de Marneffe, M. C., Descampe, A., Escouflaire, L., Fairon, C. \& Standaert, F. X. (2023). Sensibilité des explications à l’aléa des grands modèles de langage: le cas de la classification de textes journalistiques., published in \textit{Traitement Automatique des Langues, 64}.}

\maketitlepage

\section{Introduction}\label{subsec:contexte}

Large language models (LLMs) that rely on the transformer architecture, such as BERT~\cite{DBLP:conf/naacl/DevlinCLT19} and GPT~\cite{DBLP:conf/nips/BrownMRSKDNSSAA20}, have shown impressive performance in a variety of natural language processing (NLP) tasks, such as automated text classification~\cite{DBLP:journals/air/AcheampongNC21}. However, the lack of explainability of these complex models (sometimes referred to as `black-box models') is a major concern in many contexts in which they are used, especially when their decisions have important implications, for example in the legal domain~\cite{DBLP:journals/csur/ZiniA23}. Furthermore, the definition of the necessary conditions for model explainability is not yet the subject of a broad consensus~\cite{DBLP:journals/corr/abs-1901-04592}. As detailed in a recent study~\cite{DBLP:journals/corr/abs-2209-11326}, various supposedly desirable criteria for the explainability of a model have been introduced in the literature, but the relationship between these criteria is not formally established and their rigorous evaluation is often difficult.

\smallskip

The two criteria most commonly cited as fundamental to the explainability of a language model are faithfulness and plausibility. Faithfulness is defined as the ability of an explanation to accurately reflect the (algorithmic) reasoning process that led to a prediction~\cite{DBLP:conf/kdd/Ribeiro0G16,DBLP:conf/acl/JacoviG20}. Plausibility is defined as the ability of an explanation to be understandable and convincing to a human reader~\cite{DBLP:journals/corr/abs-1711-07414,DBLP:conf/acl/JacoviG20}. Various methods of model explanation have been proposed in the literature to combine these two criteria. A recent example, which we use here, is the Layer-wise Relevance Propagation (LRP) method~\cite{DBLP:conf/cvpr/CheferGW21}. It produces explanations in the form of attention maps, which are assumed to be easily understandable by a human reader~\cite{DBLP:conf/acl/SenHYKR20}.

\smallskip

A more technical (and therefore easier to quantify) criterion of the explanations of a language model is its sensitivity to different types of variation~\cite{DBLP:conf/icml/SundararajanTY17,DBLP:conf/nips/AdebayoGMGHK18}.
For example, sensitivity to input data implies that an explanation must (or must not) depend on variations in the texts being processed if these variations change (or do not change) the model's prediction.
More directly related to our concerns, it has also been proposed that an explanation should be model-sensitive, i.e.\ that it should (or should not) depend on changes in the model that affect (or do not affect) predictions.
More specifically, the concept of implementation invariance formalizes the fact that functionally equivalent models should have identical explanations~\cite{DBLP:conf/icml/SundararajanTY17}~\footnote{Functionally equivalent models have identical predictions for any input}. However, this need is mitigated by some authors, since there is nothing to prevent two different algorithmic methods from deterministically reaching the same solution ~\cite{DBLP:journals/corr/abs-2209-11326}. 

\smallskip

To the best of our knowledge, the faithfulness and plausibility of a model's explanations have so far been studied for fixed models, resulting from a given execution of their optimization. In addition, their sensitivity has been studied across models with different implementations, e.g.\  modified by removing certain features. On the other hand, the sensitivity of the explanation methods to the hyperparameters used during (different executions of) the optimization of the same model has not been systematically studied, despite the fact that the training phase of many learning methods uses hyperparameters that are chosen randomly or, at best, heuristically. In this paper, we therefore focus on learning methods which include random elements in their training process and for which the impact on optimizing the predictions' accuracy cannot be determined \textit{a priori}. It is therefore the learning process that enables the impact of these random elements to be determined \textit{a posteriori}.

\smallskip

More specifically, training a learning method usually requires the selection of a number of hyperparameters, both for the model itself (e.g.\ network size and topology) and for the optimization algorithm used (e.g.\ learning speed and batch size). In addition, stochastic optimization methods exploit a degree of randomness that is often generated deterministically from an additional parameter, usually called `seed'. The seed is then used to initialize a pseudo-random number generator to produce the amount of randomness required by the optimization algorithm. This parameter is usually made public for the purpose of reproducibility of results. However, there is nothing preventing it from being used as a hyperparameter nor preventing the comparison of the quality of the models obtained with different seeds. Considering the seed as a hyperparameter in training is rarely recommended in practice, as it does not allow any strategy other than an exhaustive search (as nothing distinguishes the randomness generated with one seed from that generated with another).
Nevertheless, the seed is of interest to us in this context because it constitutes a pristine example of a completely random element used during model training. Here we examine the effects of random seeds used as hyperparameters in the training phase. 

\section{Research question and contributions}\label{subsec:contrib}

This paper thus focuses on whether the sensitivity of large language models to the random elements of their training can be significant enough to affect their explainability.

\smallskip

Obviously, since a training process with several randomly chosen hyperparameters can lead to different models, these models can have different classification accuracy. The hyperparameter values that lead to the best performing model are generally selected. Our study therefore requires a first restriction: we are interested in the subsets of hyperparameters that lead to models with sufficiently close accuracy. We define (statistically) equivalent models as models whose differences in accuracy are not statistically significant. Moreover, even statistically equivalent models are not necessarily functionally equivalent: we are interested in the subset of data inputs for which these models output the same prediction. We refer to these equivalent model inputs giving the same prediction as `concordant'.\footnote{These notions of equivalence and concordance could be defined for other performance metrics without altering the general conclusions of the paper.} 

\smallskip

Informally, these definitions allow us to restrict our study to subsets of inputs for which there is no way to determine whether one model is preferable to another. In this context, we claim that if a learning method leads to a non-negligible set of equivalent models whose explanations differ, then limiting ourselves to the explanation of a single model obtained with that learning method is insufficient. In particular, if it is true that different algorithmic methods can lead to the same solution (in which case the explanation of a single method may be sufficient),
the presence of randomness in the process of training equivalent models forces us to ensure that the statistical distribution of the explanations from equivalent models is sufficiently different from the uniform distribution. Such a distribution would imply that all possible explanations are equiprobable, and the choice of an explanation over another would then be arbitrary.

\smallskip

In the first section of the paper, we show that for a given reasonable combination of a learning method and explanation tool, such non-trivial sets of equivalent models can be observed in practice. For this purpose, we use the LRP method mentioned above and explain the results of several transformer models,
all based on a fine-tuned CamemBERT model~\cite{DBLP:conf/acl/MartinMSDRCSS20} with the same training set and different randomly chosen seeds. This combination of a language model, a fine-tuning method and an explanation tool is applied to a specific text classification task in French. This task consists in predicting whether French press articles belong to the journalistic genre of opinion (editorials, chronicles) or news (dispatches, newswire). This sub-task of the opinion mining field is considered particularly complex~\cite{DBLP:journals/kbs/KumarR15}. It is however becoming increasingly important in the new modern information ecosystem as the polarization of social networks increases.

\smallskip

We emphasize (and will return to in our conclusions) that our claim is limited to the observation that, in the presence of explanations which depend on random factors, it is necessary to characterize this randomness and that this characterization can influence certain desirable criteria of a model's explanations. As a first step in this direction, we propose a visual characterization based on box plots. These plots highlight that the randomness of the learning process has a negative impact on the minimality of explanations, which is sometimes presented as an additional desirable criterion~\cite{DBLP:journals/ai/Miller19}. Following Occam's razor, this criterion suggests that among different explanations, the simplest one is usually the best. We also emphasize that our statement is limited to a reasonable combination of a learning method and an explanatory tool applied to a specific task. It is therefore possible that other combinations could be used to reduce the sensitivity to randomness, or that other tasks are inherently less prone to it. Finally, while we claim that characterizing the sensitivity to randomness of equivalent model decisions is a necessary condition for their explainability, we do not claim that the effect of this sensitivity is positive or negative for other desirable criteria of explanations of these decisions, such as their plausibility. These clarifications are also discussed in the conclusions of this paper.

\smallskip

Given the sensitivity of the explanations provided using CamemBERT, 
we also explore a more traditional NLP method based on textual features, which we use to train a logistic regression model.
Such classification methods are usually recognized as being easier to explain~\cite{DBLP:conf/fire/GemesKRR21}. However, they are limited by achieving lower accuracy for many applications and are therefore often neglected in favor of methods using deep learning~\cite{DBLP:journals/corr/abs-2008-00364}. They represent a very different trade-off between accuracy and 
explainability. We use linguistic attention maps to visualize the explanations of a feature-based model in a format similar to that of LRP, assigning a certain relevance to each token (word or punctuation mark) in a text classified by the model. Unsurprisingly, we find that this type of textual feature-based model produces predictions with (slightly) reduced accuracy, but that its training converges on a unique solution producing identical explanations for a given text.

\smallskip

Based on this observation, and although several recent studies have proposed the insertion of theoretical features into transformer models to improve their potential for explainability \cite{DBLP:conf/acl-alw/KoufakouPBP20,DBLP:journals/ipm/PolignanoBBGVB22}, we propose, conversely, to enrich our model based on linguistic features with a set of new features extracted from explanations derived from a transformer model. Using this approach, we show that it is possible to improve the accuracy of the linguistic model (which is still lower than that of the transformer models) while retaining deterministic results and thus invariant explanations for a given prediction.
At the very least, this hybrid approach suggests the value of large language models in exploratory tasks, such as identifying working hypotheses to be confirmed by inductive analysis. On the other hand, it leaves open the fundamental problem of the explainability of large language models.

\smallskip

Finally, we mention complementary work investigating the sensitivity of explanations to the hyperparameters (chosen randomly or heuristically) used in the explanation methods themselves~\cite{DBLP:conf/cvpr/BansalAN20}. This sensitivity is generally presented as detrimental because it implies unpredictability of explanations for a given model and prediction. However, this paper differ from ours, which focuses on the randomness of the model's hyperparameters used during training rather than on the explanation methods. We also mention the recent paper published by \citeasnoun{DBLP:journals/corr/abs-2210-13393}, which highlights different types of use (justified or hazardous) of randomness in training. However, this general discussion is not directly related to the issue of explainability.

\section{State of the art}\label{sec:sota}

\subsection{Explanation methods}\label{subsec:methex}

For text classification models, established explanation methods fall into two categories, based on their scope~\cite{DBLP:conf/ijcnlp/DanilevskyQAKKS20}. Global explanation methods aim to explain the reasoning of the model for classifying any document. On the other hand, local explanation methods focus on the reasoning of the model for a given prediction. Local explanation methods (which are the most relevant to our investigations) are further divided into several subcategories, depending on whether they are based on similarity with other examples, analysis of the internal structure of models, back-propagation mechanisms, or counterfactual analysis~\cite{DBLP:journals/corr/abs-2209-11326}.
In this paper, we focus on methods based on back-propagation mechanisms that attempt to interpret the attention layers used by transformer models~\cite{DBLP:conf/emnlp/KovalevaRRR19,DBLP:conf/blackboxnlp/ClarkKLM19}. Although the debate about whether attention alone can be used as a valid source of explanation remains open~\cite{DBLP:conf/acl/BibalCAWWFW22}, recent work has shown that combining multiple attention layers (across model gradients) can generate convincing explanations~\cite{DBLP:conf/nips/SrinivasF19,DBLP:conf/acl/AbnarZ20}. 
Among these, we use the explanation method based on layer-wise relevance propagation (LRP). Despite the inherent shortcomings of back-propagation explanation, LRP appears to be one of the most faithful  methods~\cite{DBLP:conf/wassa/ArrasMMS17} and therefore provides a good starting point for analyzing the sensitivity of transformer model explanations to randomness.

In the context of text classification, the LRP method explains the predictions made by models based on deep learning by evaluating the importance of each token in the text. This importance is measured by tracking, layer by layer, the contribution of the token to the prediction outputted by the model~\cite{bach2015}. This process assigns a relevance score to each token, starting from the output value and back-propagating via conservation constraints.~\footnote{To obtain a value per token, readable in natural language, and not per BERT `WordPiece' (as originally encoded by the CamemBERT architecture), we concatenate the different parts and compute the average of the sum of their attention values.} Different rules define how the relevance of a token at one level of the model should be propagated to the previous level, with the constraint that the relevance scores at each level must sum to 1 to reach the final prediction. As the propagation constraint is more difficult to satisfy for certain layers, this method is continuously being improved~\cite{DBLP:conf/icann/BinderMLMS16,DBLP:conf/acl/VoitaTMST19}. The resulting explanations are usually considered more faithful to the reasoning of the model than other explanation methods, such as those based on perturbation~\cite{DBLP:conf/wassa/ArrasMMS17}.

\subsection{Visualization of explanations}\label{subsec:visual}

The way in which explanations are visualized may influence their plausibility to human readers~\cite{DBLP:conf/nips/ReifYWVCPK19}. One of the most prevalent approaches to visualizing the explanations of predictions made by a text classification model is the use of attention maps~\cite{DBLP:conf/naacl/LiCHJ16}. These maps comprise the highlighting of text elements which were identified as the most influential in the model's decision-making process, employing varying shades of color. Consequently, attention maps are constrained to providing localized explanations at token-level. They are thus unable to highlight the influence of other types of features that may be decisive in the model's prediction, such as long-term dependencies and syntactic or semantic relationships between different explanatory elements. The attention map format nevertheless has the advantage of being very comprehensible a priori to human readers, which contributes to the plausibility of the explanations (this format allows the importance of each token to be visualized separately, through token-level attention). In this article, we use the term `attention' to refer to the importance attributed to a given token in an explanation produced by any method for any model, regardless of whether the model is based on the attention mechanism or not.

\subsection{Subjectivity in journalism}\label{subsec:subjectivity}

The concept of objectivity has been a central topic of debate in journalism for decades~\cite{schudson2001}. Since the end of the twentieth century, objectivity has been regarded as one of the most crucial values of the profession. Many journalists view objectivity as an ideal towards which they should strive, despite recognizing that total journalistic objectivity is unattainable~\cite{lagneau2002}. The subjectivity inherent in the journalistic process is linked to the inevitable operations of selection and decision-making that permeate every stage of the editorial process of transmitting information. These include choosing a story, deciding on its format, giving priority to certain articles over others, and many other subjective decisions~\cite{tong2021}.
Many choices are also made by the journalist when writing the article. These include framing subjects, ordering quotations, and choosing which words to use over others. The presentation of facts is always influenced by the author's personal interpretation of those facts, guided by their point of view and experiences. This makes the quest for objectivity in news reporting particularly complex~\cite{munoz2012}.

Consequently, journalists are taught to utilize a range of stylistic techniques in order to appear as objective as possible in their articles, in accordance with the `strategic ritual of objectivity' proposed by \citeasnoun{tuchman1972}. This ritual involves the application of various mechanisms to neutralize subjectivity, which serve to attenuate or conceal the influence of the journalist's opinions on the text of the article \cite{koren2004}. Such recommendations, which are taught in journalism textbooks, imposed in newsrooms or corrected by proofreaders, include the systematic citation of sources of information, the use of impersonal sentences and neutral lexis, and the absence of figurative language in texts~\cite{charaudeau2006}. However, this pursuit of textual objectivity is limited to articles belonging to the genres of news, such as press agency dispatches, and does not extend to the opinion genres of journalism, such as editorials or columns \cite{grosse2001}.

\subsection{Classification of opinionated articles in NLP}\label{subsec:opdet}

In NLP, the writing techniques employed by journalists to convey a sense of objectivity in their articles can be employed to classify texts according to their genre, namely news or opinion. \citeasnoun{DBLP:journals/coling/WiebeWBBM04} consider linguistic subjectivity to be a continuum, with objective sentences being those devoid of significant expressions of subjectivity. The aim of this work is to identify potentially subjective elements in English texts. Texts containing few of these subjective elements are considered non-subjective or neutral ~\cite{DBLP:conf/aaai/RiloffWP05}. Over the years, several markers of subjectivity in press articles in different languages have been analyzed and evaluated using different approaches. \citeasnoun{DBLP:journals/nle/KrugerLSWS17} employed a series of twenty-eight linguistic traits, including lexical complexity and the number of figures present in the text, to classify opinion and news articles published by American newspapers. Their findings demonstrated the predictive power of certain traits for this classification task. A similar study was conducted on a corpus of French articles, in which thirty features were evaluated and nineteen of them were combined to build a classification model for opinion and news texts~\cite{escouflaire2022}. 

In recent years, traditional approaches based on linguistic features have been largely replaced with large language models based on transformers~\cite{DBLP:conf/nips/VaswaniSPUJGKP17}, as exemplified by the architecture of RoBERTa~\cite{DBLP:journals/corr/abs-1907-11692}, which serves as the foundation for the CamemBERT model~\cite{DBLP:conf/acl/MartinMSDRCSS20}. The latter is a transformer model, trained on a corpus of French texts, which has been demonstrated to outperform previous methods in a range of NLP tasks, including text classification tasks~\cite{DBLP:conf/taln/BaillyBG21,chenais2021}. However, models such as CamemBERT require significantly more computational resources than traditional feature-based models~\cite{DBLP:journals/ipm/CunhaMGCRNVFMAR21} and have greater architectural complexity. Transformer models have been successfully used for a range of tasks within the journalistic domain, including the detection of fake news~\cite{DBLP:journals/nms/VargoGA18,DBLP:conf/nips/ZellersHRBFRC19}. However, to the best of our knowledge, no studies have yet been conducted on the classification of opinion and news articles using transformer models, particularly in French.

\section{Methods}\label{sec:methodo}

\subsection{Data}\label{subsec:corpus}

We use the {\og}RTBF-InfOpinion{\fg} corpus of \citeasnoun{bogaert2023}, which contains 10\,000 French press articles published between 2012 and 2021 on the website of RTBF (Radio-Télévision Belge Francophone, \url{www.rtbf.be}), the French-speaking Belgian public service media. This corpus was extracted from the RTBF open-access corpus~\cite{escouflaire2023}, by selecting 5,000 articles identified as opinion pieces by their authors or by the media, and 5,000 news articles attributed to specific categories on the RTBF website ({\og}Belgium{\fg}, {\og}World{\fg} and {\og}Society{\fg}) dealing with subjects similar to those discussed in the opinion articles. The RTBF-InfOpinion corpus is therefore divided into two balanced classes: \textit{news} and \textit{opinion}. It contains a total of 5,323,166 tokens. On average, opinion articles contain 705 tokens, against 360 for news articles. We divided the RTBF-InfOpinion corpus into training (80~\%), validation (10~\%) and test (10~\%) sets, all balanced between the two classes.

\smallskip

In order to assess the robustness of the models to changes in the data, a second corpus was created. It consists of press articles published by another medium, \textit{Le Soir} (\url{www.lesoir.be}), which is the most popular daily newspaper in French-speaking Belgium. In this work, the LeSoir-InfOpinion corpus serves only as a test set for the classification models. It was built following the same methodology and preprocessing steps presented by \citeasnoun{bogaert2023}. This corpus contains 1,000 articles published online between 2015 and 2021. As with the RTBF-InfOpinion dataset, the LeSoir-InfOpinion corpus is divided equally between opinion and news articles, following the same editorial categories as those chosen for the RTBF corpus. The corpus comprises a total of 669,154 tokens, with an average of 859 tokens for opinion articles and 480 for news articles. It is available on request.

\subsection{Fine-tuned transformer model}\label{subsec:transformer}

The initial model used in the experiments is the CamemBERT model, in its basic and case-insensitive version, which contains 110 million parameters \cite{DBLP:conf/acl/MartinMSDRCSS20}. CamemBERT is based on the architecture of the RoBERTa model \cite{DBLP:journals/corr/abs-1907-11692} and was pre-trained on the French part of the OSCAR corpus (138 GB of text).
CamemBERT was selected over FlauBERT \cite{DBLP:conf/lrec/LeVFSCLACBS20}, another French transformer model, due to its architectural compatibility with the LRP explanation method \cite{DBLP:conf/cvpr/CheferGW21}, which we utilize to generate explanations. For the purpose of this work, we fine-tune CamemBERT for our task. The fine-tuning process involves training a classification head comprising a dense layer and a layer for recovering two output values, which are associated with the model's prediction. These layers are activated by a hyperbolic tangent function and include a dropout mechanism.

\smallskip

The hyperparameters influencing the fine-tuning of the model are: the learning rate, the batch size and the number of epochs. A series of combinations of values for each of these hyperparameters were empirically evaluated, with the optimal values selected according to the accuracy obtained by the model on the validation set and for the random seed 0. For the learning rate, we tested values ranging from $1 \times 10^{-6}$ to $1 \times 10^{-4}$ and selected the value $2 \times 10^{-5}$. For the batch size, we tested values ranging from 1 to 64 and selected the value 4. In relation to the number of epochs, we evaluated the accuracy obtained by training the model for 1 to 4 epochs, with the optimal number being 2 epochs. 

\smallskip

The random elements of the CamemBERT model training that we are studying relate exclusively to fine-tuning. They are governed by a random seed used for optimization, which influences three key aspects: (\textit{i}) the initialization of the weights of the classification head, (\textit{ii}) the order of the texts in the training set, and (\textit{iii}) the neurons targeted by the drop-out technique aimed at limiting overfitting. We did not modify this third parameter: it remained at its default value (10~\%).
We fine-tuned the model over two epochs for the news vs. opinion classification task using the RTBF-InfOpinion training set. The accuracy of the model was evaluated at each epoch on the validation set and at the end of the fine-tuning on the two test sets (RTBF- and LeSoir-InfOpinion). 

\subsection{Feature-based model}\label{subsec:LR}

The second model used in the experiments is a classifier utilizing nineteen textual features derived from the state of the art on linguistic subjectivity and identified as effective predictors of opinion in French-language press discourse~\cite{escouflaire2022}.
Most of these features are based on the presence or proportion of specific tokens or types of tokens in the text to be classified. The following linguistic features are considered: adjectives, verbs, first person pronouns and determiners, relative pronouns, the indefinite pronoun `on' (\textit{we}/\textit{one}), expressive punctuation marks (semi-colons, exclamation marks and question marks), inverted commas, numbers, negation words, words longer than seven characters, words appearing in the sentiment lexicon of \citeasnoun{new2004} or the NRC lexicon~\cite{DBLP:journals/ci/MohammadT13}. Two features are not related to the tokens but apply to the entire article: Carroll's corrected type-token ratio~\cite{carroll1964} and the average word length.
The classifier uses a binomial logistic regression. We used a grid search to identify the optimal combination of hyperparameters for the regression. The resulting model will be referred to as LING-LR, an abbreviation derived from the linguistic traits upon which it is based. In this paper, it will serve as an example of a model always provinding stable explanations, which we will seek to enrich in Section~\ref{sec:exp_sens}.

\subsection{Explanation methods}\label{subsec:meth_exp}

\begin{figure}[ht]
    \begin{center}        \includegraphics[width=1\columnwidth]{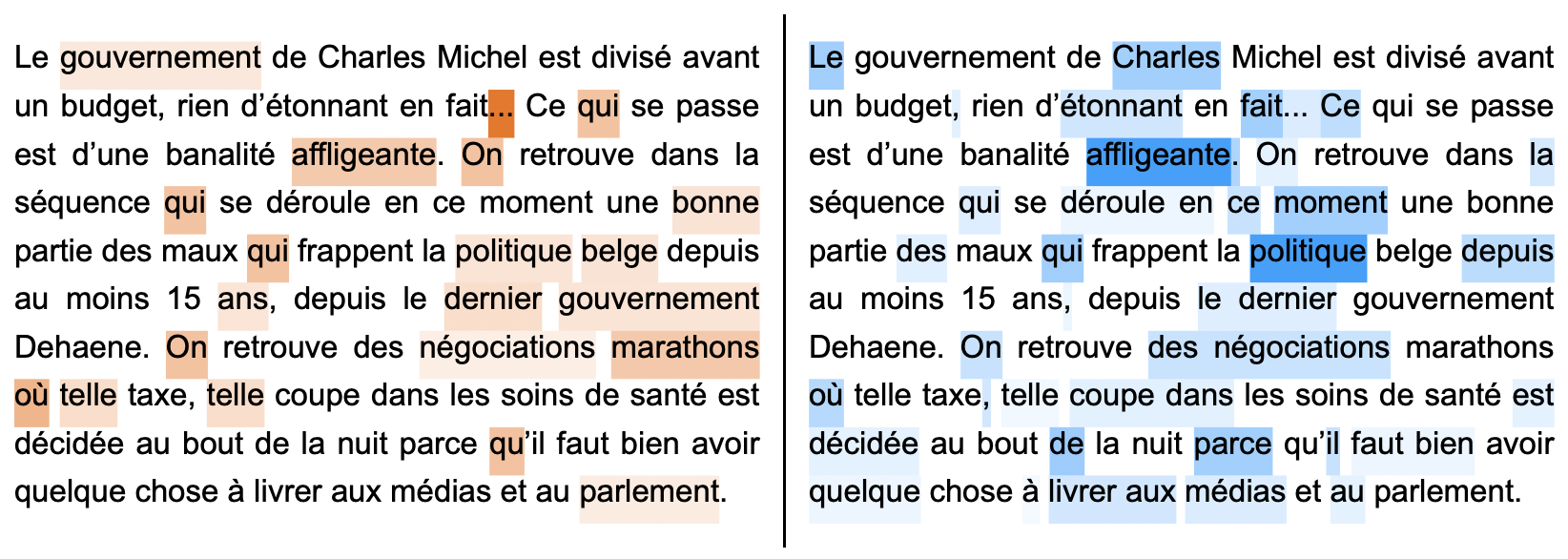}
    \end{center}\vspace*{-0.5cm}
    \caption{Attention maps derived from the linguistic model (left, in orange) and from CamemBERT (right, in blue) for the same article from the validation corpus. Both models correctly predict the article in the \textit{opinion} class.}
    \label{exempleMaps}\vspace*{-0.2cm}
\end{figure}

To generate and visualize explanations of the predictions made by the classifiers, we choose the format of token-level attention maps. This format allows us to produce explanations that are plausible for human readers~\cite{DBLP:conf/acl/SenHYKR20}. For each model, we use an explanation method to produce explanations at token-level in the form of attention maps: layer-by-layer relevance propagation (LRP), which seeks to identify the attention deployed by the fine-tuned CamemBERT \textit{transformer} model, and a `linguistic attention map' method (CAL) for the feature-based model (LING-LR). Illustrations of both types of attention maps are presented in Figure~\ref{exempleMaps}.

\subsubsection{Layer-wise relevance propagation (LRP)}\label{subsubsec:LRP}

In order to derive explanations from the predictions of the CamemBERT model, we use the LRP method relying on backpropagation, described in Section~\ref{subsec:methex}. This method computes an attention value for each token of a text based on its influence on the prediction made by the explained model. It allows to visualize the explanation in the shape of an attention map. We use the version of LRP developed by \citeasnoun{DBLP:conf/cvpr/CheferGW21} for the BERT interface, which we adapted to make it compatible with the RoBERTa interface (on which CamemBERT is based).

\subsubsection{Linguistic attention maps (LAM)}\label{subsubsec:CAL}

Linguistic attention mapping (LAM) is a method that was developed as part of this work. Its objective is to produce token-level readable explanations of the predictions of our linguistic model. They allow us to visualize the importance attributed to each token in a text classified by this model (or by any model based on textual features). The method produces an attention map that highlights the tokens that contribute to the most decisive features for the predicted class, depending on the class predicted by the model for an article. For the LING-LR model, we compute the importance of each token from the regression coefficients assigned to each features to which the token is associated. 
An token is highlighted in a darker color if it is associated with one or more linguistic features and if the weights of these features in the model are high. Thus, the linguistic attention attributed to a word \textit{i} for a feature \textit{j} can be written as:
\begin{equation}
    A_{ij} = \left\{\begin{array}{ll} 
    \frac{w_{ij}}{\sum_i{w_{ij}}} \times T_j & \mbox{if sign($T_j$) = sign(prediction)},\\
    0 & \mbox{in the opposite case},
 \end{array}\right.
\end{equation}
where the first term represents the relative importance of the word \textit{i} for the feature \textit{j} and the second represents the coefficient associated with the feature \textit{j} in the regression.\footnote{For categorical traits (adjectives, verbs\ldots), this term is 0 or $\frac{1}{n}$, but it may differ from these limiting cases when considering continuous variables (imageability, concreteness\ldots).}. Finally, the attention attributed to a word \textit{i} for all the combined features is averaged as $ A_i = \sum_j{A_{ij}}$.

LAMs generate explanations comparable to those generated by LRP (Section~\ref{subsubsec:LRP}).
Since they view importance at token-level, they cannot reflect the importance of global linguistic features (the measurement of which does not depend on the presence of specific tokens). Since the LING-LR model contains two global features (type-token ratio and average word length), this defect is detrimental to the faithfulness of the explanations. Nevertheless, we assume that this faithfulness is at least as good as in the case of the explanations of the LRP method applied to the CamemBERT models. In this case, the same problem (explaining a model with a method that cannot reflect all its complexity) is more pronounced.

\section{Sensitivity of explanations to randomness in training}\label{sec:exp_sens}

In this section, we investigate the sensitivity of the explanations of the fine-tuned CamemBERT models to random elements in its training. To do this, we first demonstrate the existence of non-negligible sets of equivalent models in section~\ref{subsec:closeness}. We then examine the correlation between explanations obtained for the same text with models differing only in terms of the randomness used to optimize them in section~\ref{subsec:corr}.
We continue with a visual characterization of the effect of this randomness on explanations in section~\ref{subsec:boxplot}. Finally, we offer a qualitative analysis of specific attention maps to illustrate the previous assessments. We also compare the results obtained with those of the LING-LR model, which may help clarifying our discussions.

\subsection{Generation of statistically equivalent models}\label{subsec:closeness}

To generate equivalent models, we first repeated the fine-tuning of the CamemBERT model 200 times (with 200 different random seeds), using the same set of 8\,000 texts during each training. We then estimated the accuracy on the test set for all models and for subsets of models, selecting the models that produced the closest or highest accuracies. Finally, we computed a $\epsilon$ parameter corresponding to the difference between the accuracy of the most accurate model (a) and the least accurate model (b) in the sets studied, to determine whether this difference was significant or not. To do this, we estimated the $Z$-statistic~\cite{LehmannZtest}, which is used to determine whether two proportions (in this case, accuracies) differ:
\begin{equation}
    z = \displaystyle\left\lvert\frac{a-b}{\sqrt{\frac{\frac{a+b}{2}*(1-\frac{a+b}{2})}{n}}}\displaystyle\right\rvert.
\end{equation}
We assumed that the differences in accuracy were significant for $z$ greater than 1.96, which corresponds to a $p$-value of less than 0.025. For lower $z$ (and higher $p$ values), we conclude that the accuracy of the models does not differ significantly. We therefore define the models in the corresponding sets as (statistically) equivalent, since their accuracy does not allow us to prefer one of them over the other.

\smallskip

\begin{table}[!h]
\small
\centering
\begin{tabular}{lcccc}
    \toprule
     & Min. Acc. & Max. Acc. &  $\epsilon$ & $p$-value \\\hline
    CamemBERT(200 models) & 93,1 & 96,6 & 3,50 & $2,8 \times 10^{-7}$\\\hline
    CamemBERT(150 closest) & 94,5 & 95,9 & 1,40 & 0,0191\\
    CamemBERT(150 most accurate) & 95,0 & 96,6 & 1,60 & 0,0860\\\hline
    CamemBERT(100 closest) & 95,0 & 95,7 & 0,70 & 0,1466\\
    CamemBERT(100 most accurate) & 95,4 & 96,6 & 1,20 & 0,3227\\\hline
    CamemBERT(50 closest)  & 95,3 & 95,6 & 0,30 & 0,3245\\
    CamemBERT(50 most accurate) & 95,7 & 96,6 & 0,90 & 0,5616\\\hline
    LING-LR & 88,9 & 88,9 & 0 & /\\
    \bottomrule
\end{tabular}
\caption{Minimum and maximum accuracy, $\epsilon$ parameter and $p$-value of sets of models.}
\label{tab:accuracies}
\end{table}

The results of these estimates are presented in Table~\ref{tab:accuracies}.\footnote{Table~\ref{tab:accuracies} also gives the accuracy of the LING-LR model estimated and tested on the same texts. The training of this model converges to a unique solution and thus the $\epsilon$ parameter is zero}. They show that from the 200 trained models, we were able to select a subset of 100 equivalent models. Since the $\epsilon$ parameter decreases trivially with the number of selected models, it only serves to illustrate how hard it is to identify equivalent models. It should be noted that this definition of equivalent models depends on the size of the test set: as its size increases, the accuracies of the models are estimated more precisely and smaller differences are considered significant. This means that more models have to be trained to select a subset of equivalent models. However, as the amount of randomness used to optimize a model is virtually unlimited, this observation does not fundamentally change the problem of sensitivity that we are discussing: it only increases the computational cost of highlighting it. Since the computational times of the CamemBERT models are generally higher than those of the LING-LR model (as shown in Table~\ref{tab:AccuracyCost}), this can become a concrete problem, especially if the explainability of such (large) models requires the generation of bigger sets of equivalent models. We further discuss this in section~\ref{subsec:discuss}.  

\begin{table}[!h]
\small
\centering
\begin{tabular}{llll}
\toprule
\textbf{Models} & LING-LR & CamemBERT \\\midrule
Pre-training & 15 min. & \multicolumn{2}{l}{13 days*} \\
Training & 0.5 sec./fit & \multicolumn{2}{l}{4 min./epoch}\\
Inference & 1.3 sec. & \multicolumn{2}{l}{9.6 sec.} \\\midrule
\textbf{Method} & LAM & {\textit{LRP}}\\\midrule
Explanation & 4 sec. & 4 min.\\
\bottomrule
\end{tabular}
\caption{Computation time of LING-LR and CamemBERT models, estimated on the 1\,000 articles from the RTBF test set, using a NVIDIA RTX A6000 GPU.}
\label{tab:AccuracyCost}\vspace*{-0.5cm}
\end{table}

\subsection{Explanation correlation}\label{subsec:corr}

Having identified sets of equivalent models, we now assess the extent to which the explanations for these models differ. To do this, and as a first informal analysis to highlight such differences, we estimate the correlation between the explanations obtained for the predictions of equivalent models on concordant inputs.\footnote{The study of nearly concordant inputs would also be possible, and would simply require examining the correlations for the \textit{news} and \textit{opinion} decisions separately.} 
Specifically, we selected two pieces of text of similar length (49 tokens for text 1, 51 tokens for text 2), classified as \textit{news} articles. We generated 100 equivalent explanations for each text and constructed two vectors, each corresponding to the concatenation of 50 explanations. We then computed the correlation between these two vectors. In the case of identical (resp. random) explanations, this correlation would be 1 (resp. 0). Finally, we repeated this operation for 10,000 different vectors to compute a \textit{bootstrap} confidence interval~\cite{DBLP:books/sp/EfronT93}.

\begin{table}[ht]
\small
\centering
\begin{tabular}{clcc}
    \toprule
    & \multicolumn{1}{c}{\multirow{2}{*}{\centering \textbf{Modèle}}} & \multicolumn{2}{c}{Pearson correlation} \\
    \cmidrule{3-4}
     & & Estimation & Intervalle de confiance bootstrap\\
     \hline
    \multirow{7}{*}{Text 1} & CamemBERT (200 models) &  0,1523 & [0,0939; 0,2098] \\
    & CamemBERT (150 closest) &  0,1448 & [0,0734; 0,2162] \\
    & CamemBERT (150 most accurate) &  0,1401 & [0,0738; 0,2065] \\
    & CamemBERT (100 closest) &  0,1259 & [0,0401; 0,2139] \\
    & CamemBERT (100 most accurate) &  0,1492 & [0,0707; 0,2268] \\
    & CamemBERT (50 closest) &  0,1176 & [0,0116; 0,2494] \\
    & CamemBERT (50 most accurate) &  0,1835 & [0,0912; 0,2803] \\\hline
    
    \multirow{7}{*}{Text 2} & CamemBERT (200 modèles) &  0,3947 & [0,3424; 0,4470] \\
    & CamemBERT (150 closest) &  0,3978 & [0,3364; 0,4589] \\
    & CamemBERT (150 most accurate) &  0,4086 & [0,3505; 0,4678] \\
    & CamemBERT (100 closest) &  0,3801 & [0,3055; 0,4536] \\
    & CamemBERT (100 most accurate) &  0,4225 & [0,3518; 0,4946] \\
    & CamemBERT (50 closest) &  0,3443 & [0,2370; 0,4519] \\
    & CamemBERT (50 most accurate) &  0,4661 & [0,3781; 0,5532] \\
    \bottomrule
\end{tabular}
\caption{Pearson correlation (with bootstrap confidence interval) between the explanations of equivalent models on concordant inputs.}
\label{tab:corr_bootstrap}\vspace*{-0.5cm}
\end{table}

\smallskip

The results of these estimates are reported in Table~\ref{tab:corr_bootstrap}.
We observe that the explanations differ significantly, regardless of whether the subsets of models are selected according to the value or the proximity of their accuracies. Interestingly, we also observe that these correlations vary depending on the text. This suggests that the dependence of the explanations on randomness is sensitive to the input data, which confirms the interest in characterizing this dependency.

\subsection{Visual characterization of the sensitivity to randomness of explanations}\label{subsec:boxplot}

To illustrate the sensitivity of the CamemBERT models, we generated box plots corresponding to the explanations of two short texts from our test set.
These give an intuition of how often the explanations of equivalent models give importance to each token.

\begin{figure}[ht]
\centering
\begin{subfigure}{.5\textwidth}
  \centering
  \includegraphics[width=0.8\linewidth]{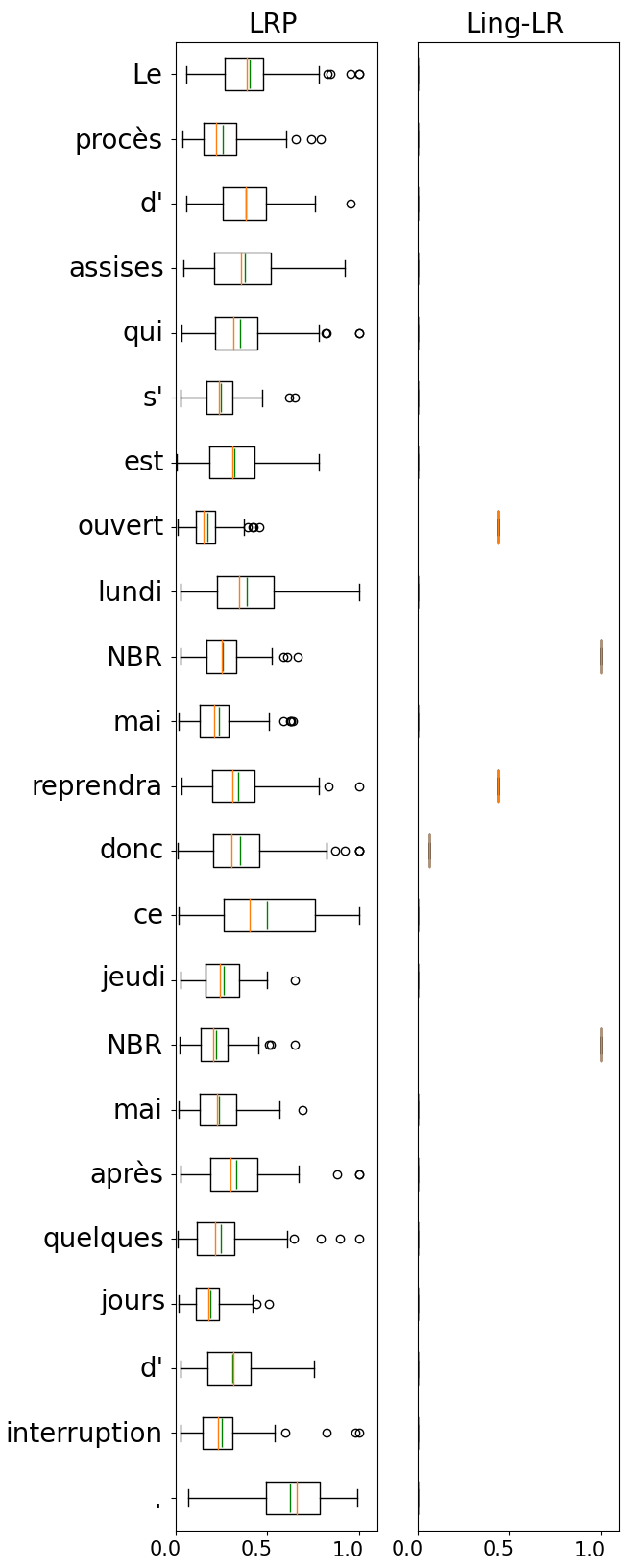}
  \caption{Explanations for text 1.}\label{fig:C100B}
\end{subfigure}%
\begin{subfigure}{.50\textwidth}
  \centering
  \includegraphics[width=0.8\linewidth]{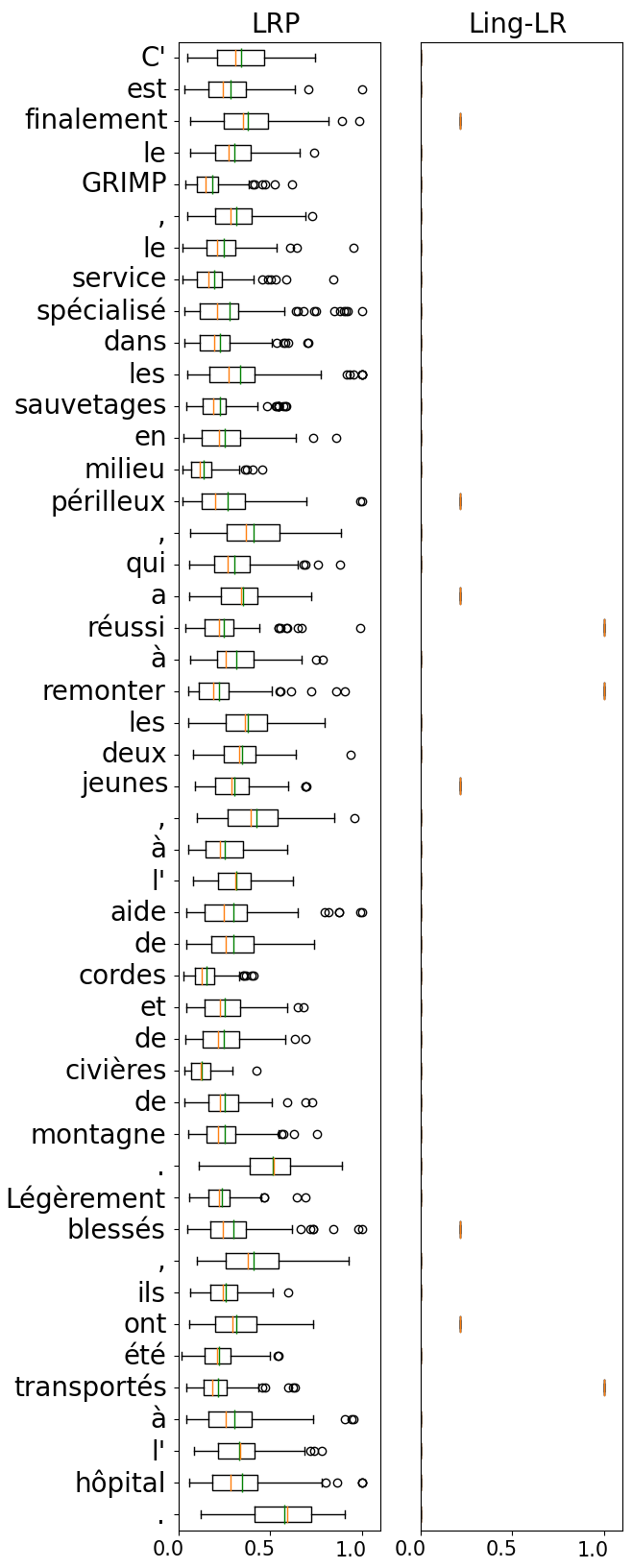}
  \caption{Explanations for text 2.}\label{fig:C50B}
\end{subfigure}

\caption{Visual characterization of the explanations of 100 equivalent models. The X axis corresponds to the distribution of token relevance.}
\label{fig:boxplot}\vspace*{-0.3cm}
\end{figure}

\smallskip

The results in Figure~\ref{fig:boxplot} show that by increasing the number of explanations, all tokens in the text end up having non-zero attention in the box plots of the CamemBERT models. Although the distribution of independently considered tokens is not uniform (and a uniform distribution of independently considered tokens does not imply a uniform distribution of all possible explanations), these results question the plausibility of the explanations obtained. In addition, and rather trivially, they also confirm a reduction in the minimality of the explanations compared to the (deterministic) explanations of the LING-LR model. This is reflected in the richer and potentially more complex distribution of explanations in the CamemBERT models.

\subsection{Qualitative analysis of chosen examples}\label{subsec:examples}

\begin{figure}[ht]
\centering
\begin{subfigure}{.4\textwidth}
  \centering
  \includegraphics[width=1\linewidth]{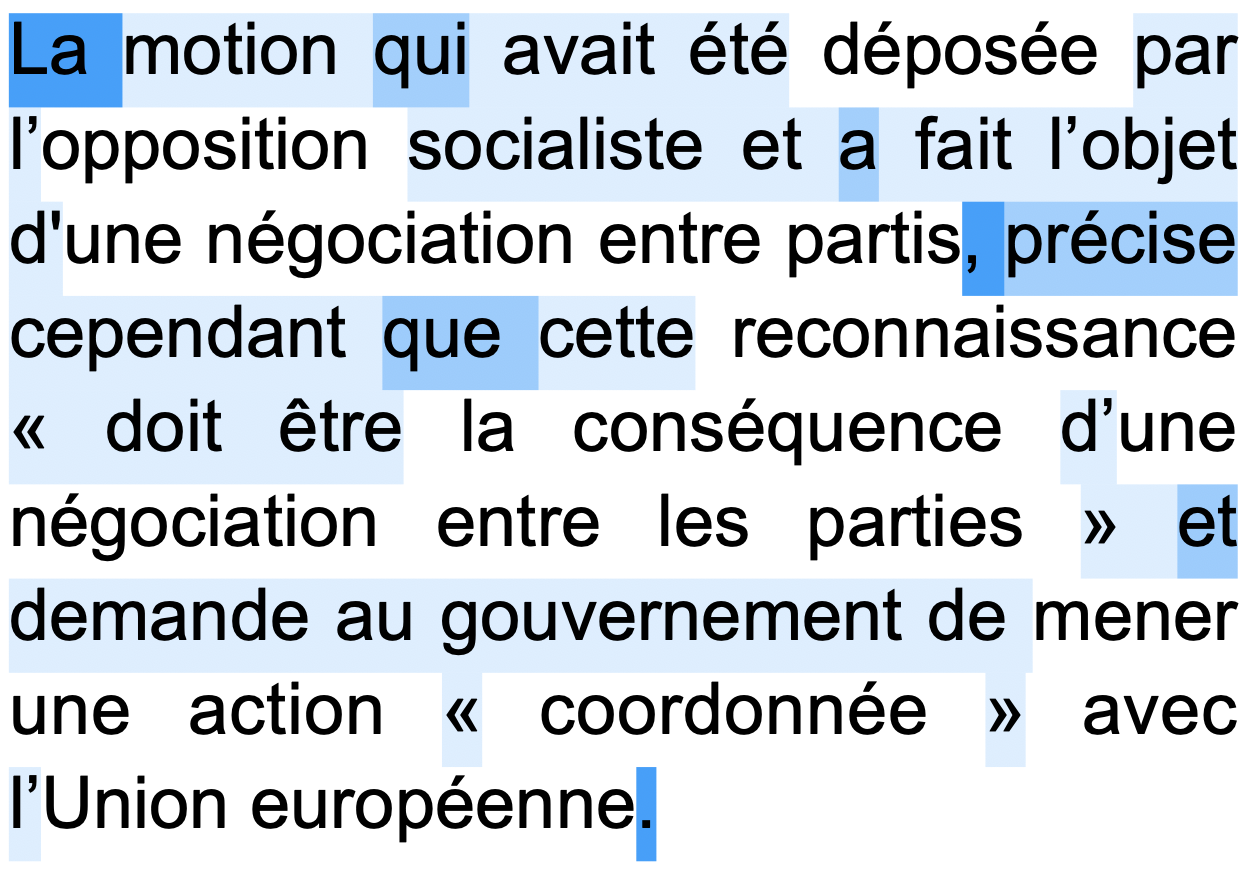}
  \caption{CamemBERT (seed = 1).}\label{fig:exp1}
\end{subfigure}%
\hspace{0.5cm}
\vspace{10pt}
\begin{subfigure}{.4\textwidth}
  \centering
  \includegraphics[width=1\linewidth]{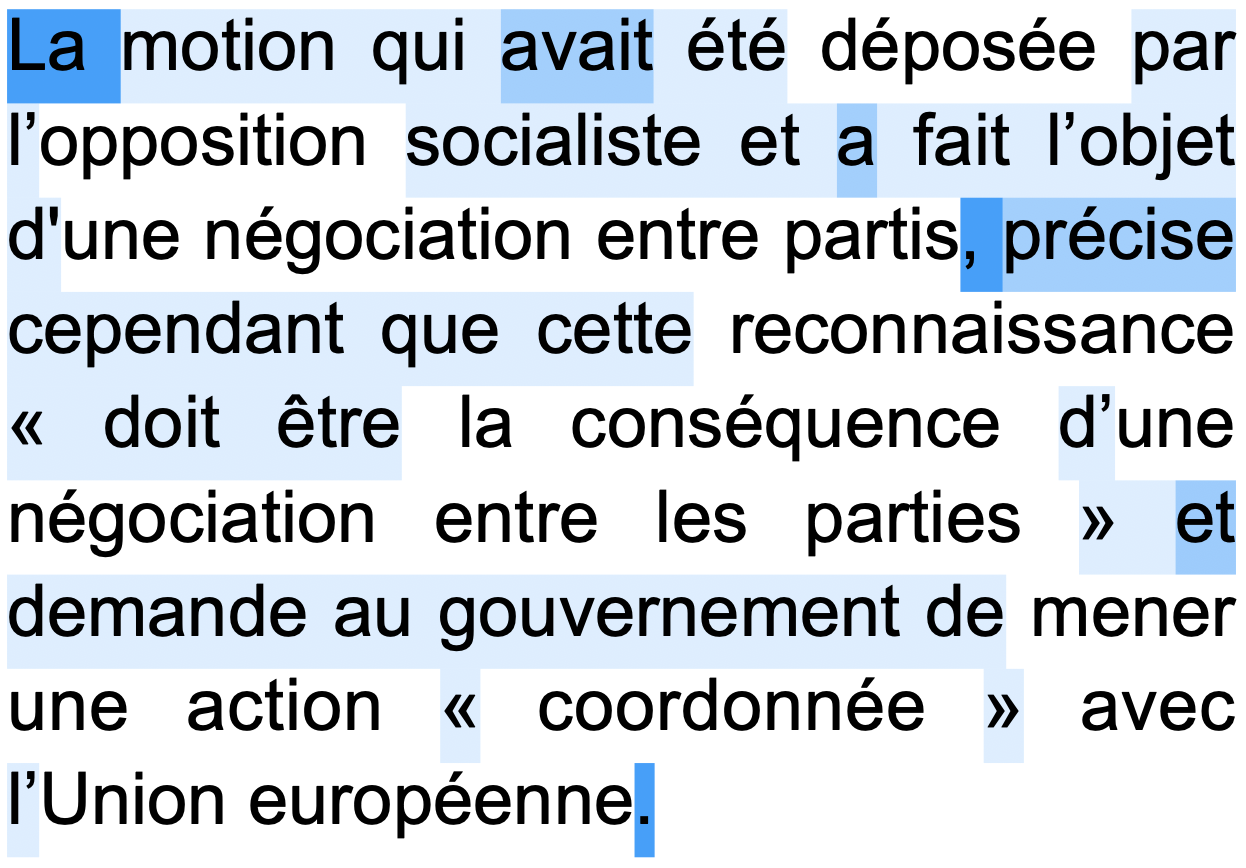}
  \caption{CamemBERT (seed = 2).}\label{fig:exp2}
\end{subfigure}
\begin{subfigure}{.4\textwidth}
  \centering
  \includegraphics[width=1\linewidth]{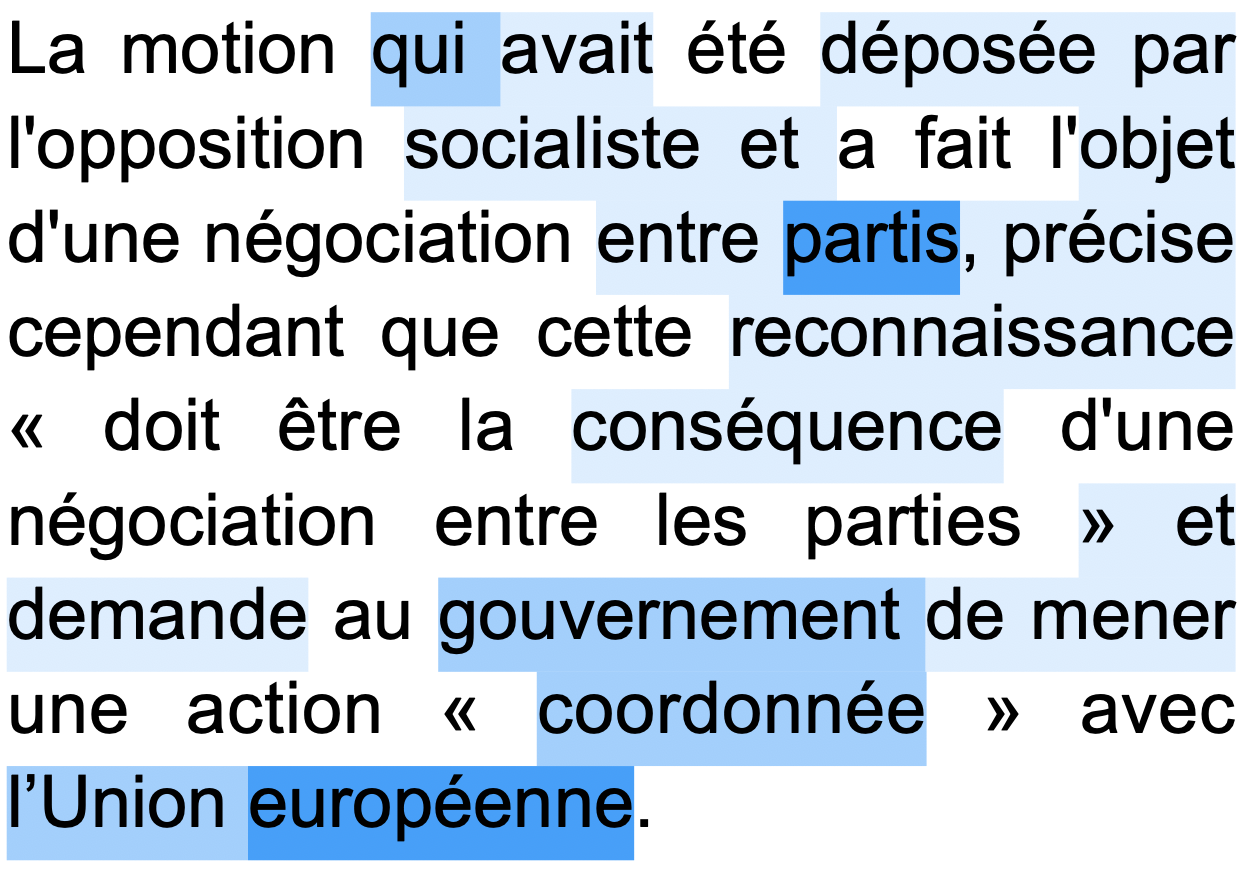}
  \caption{CamemBERT (seed = 3).}\label{fig:exp3}
\end{subfigure}%
\hspace{0.5cm}
\begin{subfigure}{.4\textwidth}
  \centering
  \includegraphics[width=1\linewidth]{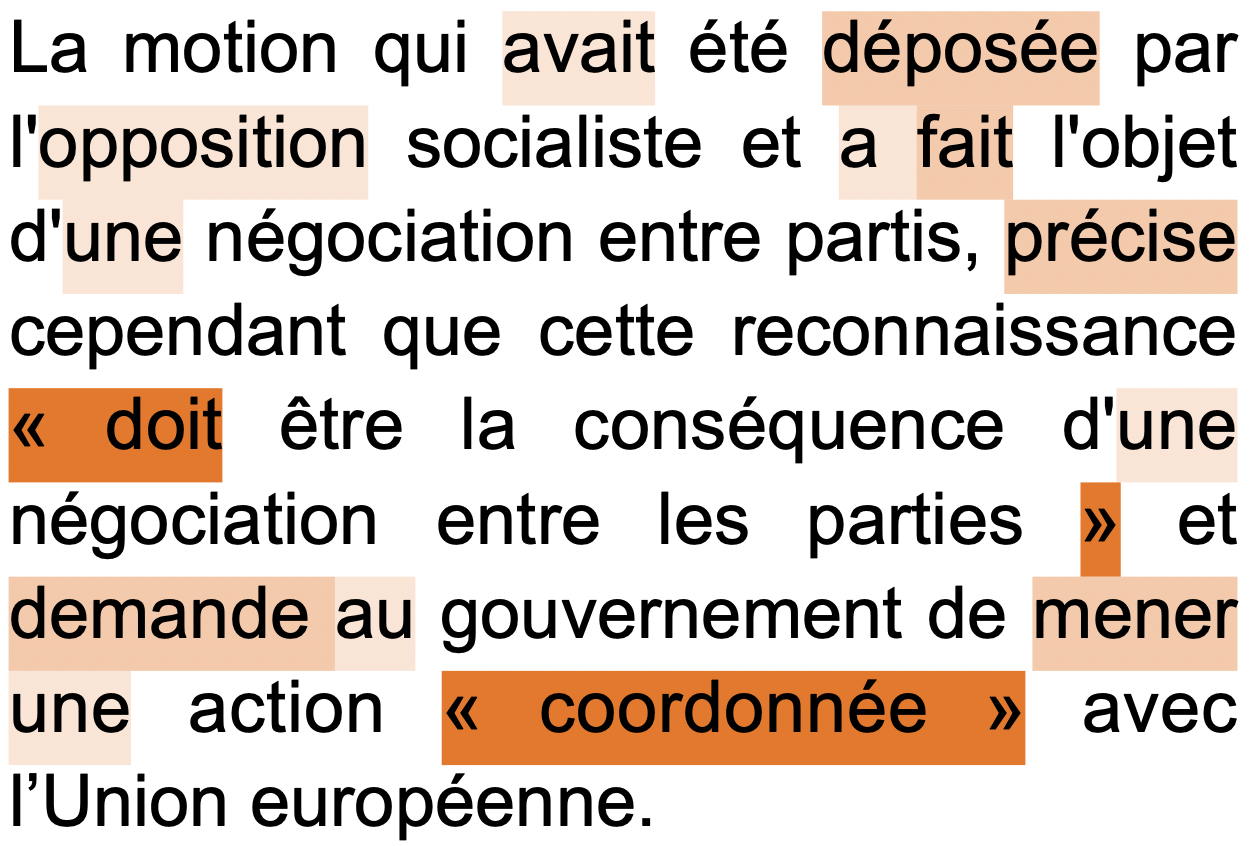}
  \caption{LING-LR.}\label{fig:exp4}
\end{subfigure}
\caption{Attention maps of four different models for a \textit{news} article (correctly classified by all models).}\vspace*{-0.3cm}
\label{fig:explanations}
\end{figure}

Finally, and to confirm that the differences in explanations quantified in section~\ref{subsec:corr} and visually characterized in section~\ref{subsec:boxplot} are not simply a combination of concordant explanations and easily filtered outliers, we conclude this section with some examples of attention maps (Figure~\ref{fig:explanations}).

\smallskip

We first observe that the explanations in Figures~\ref{fig:exp1} and~\ref{fig:exp2} are visually very similar for a human reader. Despite minor differences in the attention paid to certain tokens, both models seem to focus on the same elements, which are mainly structural elements in the text's discourse: strong punctuation marks ({\og}.{\fg}, {\og},{\fg}), conjunctions (\textit{que} (`that', \textit{et} (`and'), inverted commas, and the first words of propositions. In contrast, Figure~\ref{fig:exp3}, which was created from a third seed, shows a very different attention map from the first two. Here, the emphasis lies more on semantically charged tokens, in particular nouns representing the political actors referred to in the text (\textit{partis} (`parties'), \textit{gouvernement} (`government'), \textit{l'Union européenne} (`the European Union'). These attention maps therefore suggest that very different explanations, which are therefore difficult to filter out without further analysis, can be extracted from equivalent models. The map in Figure~\ref{fig:exp4} finally shows that the tokens which most influence the LING-LR model are the presence of quotation marks and verbs. It also differs from the explanations of the CamemBERT models.

\subsection{Discussion}\label{subsec:discuss}

The results presented in this section clearly show that the explanations of large language models can be sensitive to the randomness in their training. On the one hand, we conclude that it is necessary to characterize this sensitivity, if only to be satisfied that the distribution of explanations is sufficiently different from the uniform one. Otherwise, the choice of one explanation over another would be completely arbitrary. To the best of our knowledge, the characterization of this sensitivity has not yet been the subject of systematic studies in the literature. On the other hand, these results raise the essential question of the extent to which this sensitivity is a problem. Taking the example of legal applications, a situation might occur where an automated judgment offers to a person under trial a large number of explanations for its judgment. These explanations are quite different from the point of view of human understanding, but indistinguishable from the point of view of the algorithm. This potential situation appears incompatible with the requirement that a legal judgment should be comprehensible (and presenting the litigant with one single explanation, even if it happens to be plausible, would be at least partly arbitrary). We could also find a situation where a model produces several clusters of explanations corresponding to different but plausible interpretations of legal texts, which would be less pathological. But even in this case, it seems difficult to come up with a single explanation.

\smallskip

Furthermore, the combination of this need for characterization with the computational complexity of large language models (cf. Table~\ref{tab:AccuracyCost}) suggests that simpler models may become interesting alternatives when explainability and computational cost of predictions made by a classification model are considered important for a given application. This observation motivates the attempt to enrich this type of model in the following section. In this context, we also note that the cost of characterizing sensitivity to randomness depends on the distribution of explanations. For example, as the variance of the attention given to a word in Figure~\ref{fig:boxplot} increases, more explanations have to be generated to estimate its average attention.

\section{Enriching the feature-based model}\label{sec:enrich}

We propose to improve the accuracy of a model whose training converges to a unique solution (and thus without variability regarding several explanations extracted for a given text) by using features derived from the explanations of transformer models. Here, we add to the feature-based model (LING-LR) new features found through the explanations of the fine-tuned CamemBERT models. This hybrid model will be referred to as the enriched linguistic model (LING-LR-E).

The first step consists in measuring the average attention attributed to each token occurring at least ten times in the 1,000 attention maps generated by applying the LRP method to the predictions of equivalent CamemBERT models on the articles of the RTBF-InfOpinion test set. To illustrate the potential for enrichment with reasonable computational time, we limit the number of equivalent models used to 10. We then rank all tokens based on their average attention, in descending order. This operation is repeated for the 10 models studied. We keep only the tokens appearing among the first hundred tokens with the highest attention for at least five equivalent models out of ten. Then, we qualitatively analyze the final list of tokens in order to identify linguistic patterns which can be converted into features. This method is similar to the approach presented by \citeasnoun{DBLP:conf/naacl/ZhouRS22}, which consists in deriving information about the internal reasoning of a complex model from local explanations of its predictions. We restrict our analysis to the fifty tokens which receive the most attention for each class (\textit{opinion} and \textit{news}). The two resulting lists are presented in table \ref{listeMots}.

Our qualitative examination of the fifty tokens in the \textit{opinion} list reveals several recurring patterns: axiological words (\textit{disaster}, \textit{poor}), expressive punctuation marks ({\og}. ..{\fg}, {\og}!{\fg}), verbs of thought (\textit{imagine}, \textit{forget}), words referring to abstract concepts (\textit{ideology}, \textit{humor}), or discourse markers (\textit{in short}, \textit{of course}). Regarding \textit{news}, we can identify words referring to non-deictic temporal entities (\textit{lundi}, \textit{GMT}), verbs of quotation (\textit{precise}, \textit{affirmed}), words with a high subjective frequency (\textit{computer}, \textit{airport}), i.e.\ words perceived as frequent in everyday language (\cite{balota2001}), and words referring to sources of information (\textit{according to}, \textit{AFP}). Some of these patterns overlap with features already present in the LING-LR model, such as the number of adjectives and expressive punctuation marks (for the \textit{opinion} class), while others represent original findings, such as the presence of discourse markers and the average concreteness \cite{bonin2018} of words in the text. Finally, we note that some tokens can be considered as artifacts \cite{DBLP:conf/naacl/GururanganSLSBS18} related to the data used (\textit{parking}, \textit{Flemish}).

This analysis allows us to extract 9 new linguistic features from the patterns identified in the attention lists of Table \ref{listeMots}: the ratio of deictic temporal markers, non-deictic temporal markers, thinking verbs, quoting verbs, passive verbs, and discourse markers, as well as concreteness, imageability and the average subjective frequency of words in the text. Concreteness is measured using the lexicon of \citeasnoun{bonin2018}, while imageability and subjective frequency are measured using the lexicons of \citeasnoun{desrochers2009}.
The enrichment with these nine new features (added to the nineteen features of the original LING-LR model) gives LING-LR-E an accuracy of 89.6~\% on the RTBF-InfOpinion test set, an increase of 0.8~\% compared to LING-LR (which is not statistically significant). 
On the LeSoir-InfOpinion test set, LING-LR-E achieved an accuracy of 80.6~\% compared to 76.8~\% for LING-LR, an increase of 3.8~\% (significant according to the same Z statistic and the same $p$-value of 1.96 as in the previous sections). This shows that the elements extracted from the explanations of the CamemBERT models have contributed to making the LING-LR-E model more generalizable. 
For comparison, the best accuracy obtained by a CamemBERT model on the LeSoir-InfOpinion test set is 90.5~\% (96.6~\% on the RTBF test set).

\begin{table}[ht]
\small
\centering
\begin{tabular}{ll|ll}
    \multicolumn{2}{c}{\textbf{News}} & \multicolumn{2}{c}{\textbf{Opinion}}\\
 \toprule
précise (precises) & \textit{indiqué} (indicated) & \textit{révélations} & \textit{fed} \\
\textit{jeudi} (Thursday) & \textit{mardi} (Tuesday) & \textit{chômeurs} (unemployed) & \textit{imaginez} (imagine) \\
\textit{indique} (indicates) & \textit{expliqué} (explained) & \textit{bref} (in short) & \textit{désigne} (designates) \\
\textit{mercredi} (Wednesday) & \textit{explique} (explains) & \textit{ressemble} (resembles) & \textit{pension} \\
\textit{lundi} (Monday) & \textit{vendredi} (Friday) & \textit{fout} (does) & \textit{extension} \\
\textit{précisé} (specified) & \textit{a-t-elle} (did she) & ... & \textit{latin} \\
\textit{poursuit} (continues) & \textit{souligne} (underlines) & \textit{formateur} (instructor) & \textit{aiment} (like) \\
\textit{adaptation} & \textit{souligné} (underlined) & \textit{Hitler} & \textit{inverse} (reverse) \\
\textit{ajouté} (added) & \textit{dimanche} (Sunday) & \textit{tort} (harm) & \textit{disons} ([we] say) \\
\textit{parking} & \textit{poursuivi} (pursued) & \textit{aujourd'hui} (today) & \textit{ombre} (shadow) \\
\textit{conclu} (concluded) & \textit{AFP} & \textit{accords} (agreements) & \textit{bulle} (bubble) \\
\textit{suspect} & \textit{correctionnel} (correctional) & \textit{démontrer} (demonstrate) & \textit{mélange} (mixture) \\
\textit{ajoute} (adds) & \textit{aéroport} (airport) & \textit{politiquement} (politically) & \textit{libéral} \\
\textit{samedi} (Saturday) & \textit{affirmé} (affirmed) & \textit{désastre} (disaster) & \textit{dépit} (spite) \\
\textit{trafic} (traffic) & \textit{assuré} (assured) & \textit{flamandes} (Flemish) & \textit{chômage} (unemployment) \\
\textit{locales} (local) & \textit{affirme} (affirms) & \textit{suffisamment} (sufficiently) & \textit{correction} \\
\textit{chanteuse} (singer) & \textit{selon} (according to) & \textit{pire} (worse) & \textit{illustre} (illustrates) \\
\textit{priorités} (priorities) & \textit{températures} & \textit{retraite} (retirement) & \textit{idéeologie} (ideology) \\
\textit{déclaré} (declared) & \textit{a-t-il} (did he) & \textit{médiatique} (media-related) & \textit{immobilier} (real estate) \\
\textit{disponible} (available) & \textit{rappelé} (recalled) & \textit{voyons} ([we] see) & ; \\
\textit{février} (February) & \textit{ordinateur} (computer) & ! & \textit{oublier} (forget) \\
\textit{communiqué} (statement) & \textit{organisateurs} (organizers) & \textit{pauvre} (poor) & \textit{monétaire} (monetary) \\
\textit{blessé} (injured) & \textit{km} & \textit{certes} (of course) & \textit{utilise} (uses) \\
\textit{GMT} & \textit{pourront} (will be able) & \textit{mauvais} (bad) & \textit{impôt} (tax) \\
\textit{dépistage} (screening) & \textit{visiteurs} (visitors) & \textit{calcul} (calculation) & \textit{inutile} (useless) \\
\end{tabular}
\caption{List of the 50 tokens (from left to right then top to bottom) with the highest average attention in the explanation maps derived from the predictions made by the fine-tuned CamemBERT models for the \textit{RTBF-InfOpinion} test set.}
\label{listeMots}\vspace*{-0.3cm}
\end{table}

\section{Limitations}\label{sec:limitations}

The main contribution of this article is that it highlights a question that appears to have been under-researched at this stage (is the sensitivity of explanations to the randomness of large language models significant?), even though this question crystallizes the essential difficulty of explaining the predictions of large language models. We show that this question can arise in practice for a particular combination of a learning method and an explanation tool applied to a specific task. It follows that the generality of our observations should be extended. On the one hand, the examination of other corpora (especially in languages other than French) would be interesting, as would the study of other NLP tasks, such as the detection of `fake news', which is also recognized in the field as being particularly complex \cite{DBLP:journals/nms/VargoGA18,DBLP:conf/nips/ZellersHRBFRC19}.
On the other hand, and at technical level, the evaluation of other language models and other explanatory methods would also be necessary. We give further motivation for these various potential extensions in the conclusions that follow.

\section{Conclusions and open problems}\label{sec:conclusion}

In this paper, we have investigated the sensitivity of the explanations of large language models to the random elements of their training. In particular, we have shown that models fine-tuned on the same training set but with different random hyperparameters, although they yield similar accuracies, can provide different explanations for texts on which they give the same prediction. In other words, we have observed explanations that depend on the structure of the models generated from different random parameters, rather than on the outcome of their predictions.
Since there is no reason to prefer one model over another in this context, we conclude that restricting ourselves to the explanation of a single model is insufficient and may be arbitrary. We assert that explaining the decisions of this type of model requires a characterization of their randomness. 

\smallskip

Having found that an initial characterization of the influence of randomness on explanations of large language models using box plots reduces their minimality~\cite{DBLP:journals/ai/Miller19}, we then evaluated the extent to which a simpler model allows for more compact and more stable explanations. First, we found that, unsurprisingly, a model based on linguistic features combined with  logistic regression produces explanations that do not exhibit the sensitivity to randomness of the fine-tuned CamemBERT model, at the cost of reduced accuracy.
We also observed, thanks to the attention maps derived from the two models, that these do not rely on the same tokens to predict the same classes. We therefore tried to enrich the linguistic model by extracting new features from the attention maps computed on the fine-tuned CamemBERT model (the most accurate model) and integrating them into the logistic regression model based on linguistic features (the most stable model). This approach improved the classification accuracy of this model on two test sets, without reducing the stability of its explanations.

\smallskip

Since the enriched linguistic model still shows a lower accuracy than the fine-tuned CamemBERT model, we also conclude that large language models characterize textual features, useful for classification, that are not (and probably cannot be) integrated into the feature-based model. The fundamental problem of the explainability of the decisions made by large language models thus remains open, magnified by the dependence of these models on the randomness of training highlighted in this article. Our results therefore suggest new avenues of research aimed at identifying the origin of this sensitivity to randomness and reducing it.

\smallskip

A first avenue of research would consist in assessing the impact of the highlighted sensitivity to randomness of the explanations on their plausibility. 
Indeed, asserting that the dependence of a model's explanations on randomness must be characterized does not necessarily imply a reduction in plausibility. In particular for the opinion classification task studied, it could be observed that the variability observed reflects the variability of the explanations given by human annotators. It would therefore be interesting to set up such a human annotation experiment. Investigating the extent to which explanations of equivalent models can be grouped into clusters, as mentioned in Section~\ref{subsec:discuss}, would be particularly relevant in this perspective.

Another line of research would be to assess the extent to which this sensitivity to randomness is due to a lack of faithfulness of the explanations. In particular, we could hypothesize that methods which provide simple explanations, for example based on tokens (as those used in this paper), are not adapted to the multiplicity of features exploited by large language models, and that this mismatch between a complex model and simple explanation formats increases the sensitivity of the outputted explanations to randomness. To test this hypothesis, it would be necessary to assess the extent to which a model simpler than CamemBERT (e.g.\ with fewer parameters) appears less sensitive to randomness. It would then be useful to improve the faithfulness of the explanations by adapting the methods and formats used for model explanation to the complexity of the large language models and the randomness in their training, while raising the important question of the trade-off with their plausibility. At the same time, reducing the dependence of the training of large language models on random elements could also simplify this problem.

\bibliography{refs}

\end{document}